\documentclass[letterpaper]{article} 
\usepackage{aaai2026}  
\usepackage{times}  
\usepackage{helvet}  
\usepackage{courier}  
\usepackage[hyphens]{url}  
\usepackage{graphicx} 
\usepackage{natbib}  
\usepackage{caption} 
\usepackage{algorithm}
\usepackage{algorithmic}
\usepackage{amsmath} 
\usepackage{subcaption}

\usepackage{newfloat}
\usepackage{listings}
\DeclareCaptionStyle{ruled}{labelfont=normalfont,labelsep=colon,strut=off} 
\floatstyle{ruled}
\newfloat{listing}{tb}{lst}{}
\floatname{listing}{Listing}
\title{TopoReformer: Mitigating Adversarial Attacks Using Topological Purification \\ in OCR Models}
\author{
    Bhagyesh Kumar\equalcontrib, 
    A S Aravinthakashan\equalcontrib, 
    Akshat Satyanarayan\equalcontrib, 
    Ishaan Gakhar, 
    Ujjwal Verma
}

\affiliations {
    Manipal Institute of Technology, Manipal Academy of Higher Education Manipal - 576104, India\\    
    invi.bhagyesh@gmail.com,
    aravinthakshanmain@gmail.com,
    pkmnakshat@gmail.com,
    ishaangakhar04@gmail.com,
    ujjwal.verma@manipal.edu
}

\usepackage{bibentry}

\begin{document}

\maketitle

\begin{abstract}
Adversarially perturbed images of text can cause sophisticated OCR systems to produce misleading or incorrect transcriptions from seemingly invisible changes to humans. Some of these perturbations even survive physical capture, posing security risks to high-stakes applications such as document processing, license plate recognition, and automated compliance systems. Existing defenses, such as adversarial training, input preprocessing, or post-recognition correction, are often model-specific, computationally expensive, and affect performance on unperturbed inputs while remaining vulnerable to unseen or adaptive attacks. To address these challenges, \textit{TopoReformer} is introduced, a model-agnostic reformation pipeline that mitigates adversarial perturbations while preserving the structural integrity of text images. Topology studies properties of shapes and spaces that remain unchanged under continuous deformations, focusing on global structures such as connectivity, holes, and loops rather than exact distance. Leveraging these topological features, \textit{TopoReformer} employs a topological autoencoder to enforce manifold-level consistency in latent space and improve robustness without explicit gradient regularization. The proposed method is benchmarked on EMNIST, MNIST, against standard adversarial attacks (FGSM, PGD, Carlini–Wagner), adaptive attacks (EOT, BDPA), and an OCR-specific watermark attack (FAWA).
\end{abstract}

\begin{links}
    \link{Code}{https://github.com/invi-bhagyesh/TopoReformer}
\end{links}

\section{Introduction}

The widespread adoption of OCR in critical safety and compliance workflows, such as document automation \cite{karthikeyan2021ocr}, traffic enforcement \cite{kakani2017improved}, and enterprise data extraction \cite{bazzo2020assessing}, means that even minor transcription errors can propagate to financial loss \cite{pingili2025ai}, policy breaches, or wrongful enforcement actions at scale. This dependence on OCR has grown with the broad adoption of deep neural networks (DNNs) \cite{akhtar2018threat}, which have significantly advanced recognition accuracy and enabled deployment in many high-stakes applications. However, OCR also inherits the security vulnerabilities of DNNs \cite{li2022review}. It is highly susceptible to adversarial examples, where human imperceptible perturbations to text images can cause targeted or severe transcription errors, even surviving real-world channels such as print and scan or display and camera capture.

These developments underscore a fundamental challenge for which various defense strategies have been explored to counter this threat. In general, the domain of adversarial defense has followed four directions: preprocessing through purification or denoising modules that project inputs back into a learned data manifold \cite{meng2017magnet} \cite{hwang2019puvae}; detection of anomalous inputs \cite{he2022your}, often via autoencoder-based reconstruction errors \cite{cintas2021detecting} or distributional tests \cite{folz2020adversarial}; adversarially robust training of recognition models \cite{zhao2022adversarial}, \cite{qian2022survey}; and post-processing corrections applied to suspicious text \cite{xie2017mitigating}, \cite{imam2022ocr}. However, these approaches remain costly, model-specific, or prone to failure on unseen attacks, motivating the need for more generalizable defenses. To address these limitations, TopoReformer is proposed, which is designed as a cost-alleviating, model-agnostic framework that maintains robustness even against adaptive attacks. The proposed approach operates independently of specific OCR architectures and avoids adversarial retraining, thus ensuring scalability and ease of deployment while providing stable defense across a wide range of perturbation types.

Topology \cite{edelsbrunner2008persistent} studies the properties of shapes and spaces that remain unchanged under continuous deformations, focusing on global structures such as connectivity, holes, and loops rather than exact distances. This invariance forms the core intuition behind the proposed approach: adversarial perturbations often distort only local pixel relationships, whereas the overall topological structure of the data remains intact. By embedding this structural prior into the learning process through their loss function, topological autoencoders naturally filter out such perturbations and recover the underlying manifold of clean samples. TopoReformer, unlike denoising-based defenses that rely solely on pixel-level reconstruction \cite{meng2017magnet}, uses an additional topological loss that enforces consistency between the persistent homology of the input and its latent representation. 

TopoReformer distinguishes itself from denoising models or obfuscated gradients, which were used repeatedly in many adversarial robustness papers and shown to provide a false sense of "robustness" \cite{athalye2018obfuscated}. The reshaped manifold thus performs purification, not denoising, discarding perturbations that do not correspond to meaningful topological variations—a distinction validated experimentally under adaptive attacks. The novelty of this work lies in a topology-based purification method that addresses watermark-based OCR attacks, adaptive and classical adversarial attacks such as EOT \cite{athalye2018synthesizing}, BPDA \cite{athalye2018obfuscated}, Carlini \cite{carlini2017towards}, PGD \cite{madry2017towards}, and FGSM \cite{goodfellow2014explaining}, trained solely on unperturbed data without adversarial examples. Experimental results demonstrate that the approach effectively removes adversarial artifacts, underscoring the distinctive inductive bias underlying the method.

Although some work has explored the use of topological features in related contexts \cite{goibert2022adversarial, gebhart2017adversary}, the literature on topological autoencoders to mitigate adversarial attacks remains very limited. For example, \cite{vu2025topological} leverages topological features for logit alignment, and \cite{kuang2024defense} applies topological methods for adversarial defense. However, there appears to be little to no prior work employing topological autoencoders for end-to-end purification or defense. To the best of the authors’ knowledge, this work is the first to employ Topological Autoencoders in this setting. The main contributions of the paper are summarized as follows:

\begin{itemize}
    \item \textbf{Topological purification:} A Topological Autoencoder is employed for the purification of adversarial images, followed by a lightweight reformer trained on unperturbed data. This setup suppresses adversarial artifacts while preserving the overall structure of the inputs toward a cleaner manifold. 
    
    \item \textbf{Freeze-Flow training paradigm:} A novel training scheme that blocks direct gradients to the primary encoder and routes them through an auxiliary module, encouraging reliance on topology-consistent latents to complement the purification, yielding up to 5\% improvement in classification performance under Carlini-Wagner attacks.
    
    \item \textbf{Model Agnostic OCR defense :} A drop-in, model-agnostic pipeline for any OCR model that integrates the topology-guided purifier and Reformer, providing robustness against white-box, black-box, and adaptive attacks without requiring adversarial training.
\end{itemize}

\section{Related Works}

The following subsections frame the problem space by first motivating the choice of adversarial attacks as representative benchmarks. Next, reconstruction-based reformers (purifiers), such as denoising and masked autoencoders, are reviewed with an emphasis on their limitations. Finally, topology-preserving autoencoders are introduced, highlighting their motivation in guiding the proposed pipeline.

\subsection{Adversarial Attacks} Adversarial attacks \cite{chakraborty2021survey} are deliberate, minimal perturbations to inputs or training data crafted to induce a model to produce incorrect, unintended, or targeted outputs while remaining imperceptible to humans. In evaluation, they are typically framed as worst-case, bounded perturbations that maximize a loss (untargeted) or enforce a specific prediction (targeted) under white, gray, or black-box conditions. FGSM \cite{goodfellow2014explaining} applies a single-step gradient-sign update to reveal linearity-driven vulnerabilities, while PGD \cite{madry2017towards} extends this into iterative projected updates within a bounded norm, becoming the de facto standard for robustness evaluation and adversarial training. C\&W \cite{carlini2017towards} formulates adversarial generation as an optimization problem minimizing perturbation magnitude subject to a confidence-based misclassification constraint, producing high-confidence, low-distortion examples that have historically bypassed many defenses.

Later, it was highlighted that several defenses evaluated only against such static attacks \cite{athalye2018obfuscated}  achieved a false sense of security due to gradient obfuscation, non-differentiable, randomized, or poorly conditioned transformations that blocked adequate gradient flow. Adaptive attacks were proposed to address this, where the adversary explicitly models or approximates the defense during optimization. Two major frameworks, Backward Pass Differentiable Approximation (BPDA) \cite{athalye2018obfuscated} and Expectation over Transformation (EOT) \cite{athalye2018synthesizing}, emerged as standard. BPDA substitutes non-differentiable components with differentiable surrogates during backpropagation, while EOT estimates expected gradients over stochastic transformations. This ensures that evaluations assess actual robustness rather than artifacts of gradient masking.

\begin{figure*}[ht]
    \begin{center}
        \includegraphics[width=0.9\linewidth]{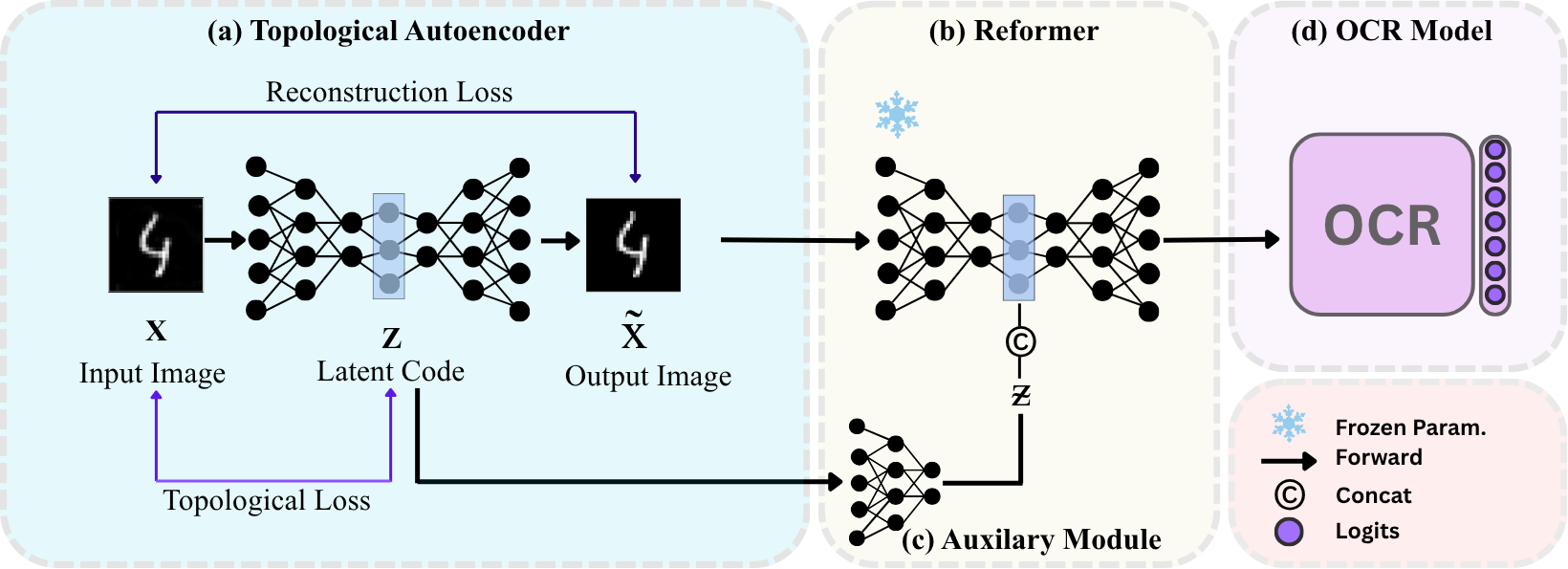}
    \end{center}
    \caption{Schematic of the TopoReformer pipeline and the freeze-flow paradigm. The Topological Autoencoder (TopoAE), Auxiliary Module, and Reformer diagrams are illustrative and do not represent the actual neural network architectures.}
    \label{fig:architecture}
\end{figure*}

\subsection{Autoencoder-based Reformers and Defenses}

Denoising autoencoder-based purifiers aim to reconstruct inputs to remove perturbations before classification. Methods such as PuVAE \cite{hwang2019puvae}, APuDAE \cite{kalaria2022towards}, and MAEP \cite{chen2025adversarial} improve performance on MNIST, CIFAR-10, and ImageNet, but often degrade accuracy on unperturbed inputs \cite{yang2024adversarial}. Earlier autoencoder-based defenses, such as \cite{meng2017magnet} and \cite{bakhti2019ddsa}, also reform threshold input, but minor L1 perturbations can bypass them \cite{lu2018limitation}. Defenses that combine denoising with randomized transformations fail under adaptive attacks using Expectation Over Transformation (EOT), which averages over randomness to circumvent purifiers. Notable examples include \cite{dhillon2018stochastic, xie2017mitigating, guo2017countering}, all of which saw their accuracy drop near 0\% under EOT. In contrast, TopoReformer allows a classifier to make robust predictions even under adaptive attacks, with the help of global manifold preservation to maintain performance on both unperturbed and perturbed data, which is an issue with most adversarial-aware training paradigms that sacrifice performance on the unperturbed or clean samples in the attempt to perform well on adversarial inputs.

\subsection{Topological Autoencoders}
Topology-preserving autoencoders \cite{moor2020topological} incorporate persistent-homology loss to retain the multiscale connectivity of inputs in latent or reconstructed spaces, preserving structural cues such as stroke and loop continuity, which are critical for optical character recognition. These autoencoders compute topological signatures for input and latent spaces, minimizing their discrepancy, producing representations that maintain global structure while achieving low reconstruction error on real image data. This provides a principled basis for structure-faithful purification without the need to retrain the classifier. Topological autoencoders have previously been applied for preserving structure representation learning \cite{sainburg2021parametric}, improving graph neural network embeddings \cite{horn2021topological}, and detecting anomalies in physics \cite{ngairangbam2025enhancinganomalydetectiontopologyaware} by capturing deviations from the learned topological manifold. No work has explored the use of topological autoencoders for adversarial defense or input purification.

\section{Methodology}

The methodology section is structured as follows: first, the Topological Autoencoder (TopoAE) is introduced along with a detailed discussion of persistent homology, persistence diagrams, and the associated topological loss. This is followed by a description of the Reformer module, the auxiliary model, the proposed training paradigm, and the visualization of the topological latent space.

\subsection{A Topology-Preserving Autoencoder}

The pipeline proposed, illustrated in Fig. \ref{fig:architecture}, consists of three main components: the topological autoencoder (TopoAE), the Reformer (VAE), and the Auxiliary module. The input image is first processed by the TopoAE, which generates: a latent representation that encodes the topological structure of the input, and a reconstructed output (purified) that preserves the essential topological properties while filtering out topologically irrelevant variations. This purified image is then passed through the Reformer, which adjusts the output of the TopoAE to better align with the expected manifold of the OCR model. Simultaneously, the latent vector is routed through the Auxiliary Module, which injects this topological information into the bottleneck of the Reformer. Finally, the decoder of the Reformer reconstructs the purified image, which is then passed to the downstream classifier for final transcription.

The central motivation behind this work lies in the working of the autoencoder based on \textit{persistent homology} \cite{zomorodian2004computing}, a computational topology tool used to capture connectivity-based structures in data. Given a simplicial complex $K$, homology identifies $d$-dimensional features such as connected components ($d=0$), loops ($d=1$), and voids ($d=2$). Instead of relying on a single complex, persistent homology tracks how these features appear and disappear across multiple scales using the Vietoris--Rips complex $R_\epsilon(X)$, built from pairwise distances within a point cloud $X$. As the scale ($\epsilon$) increases, $R_\epsilon(X)$ expands and features merge or vanish, forming \emph{persistence diagrams} $D_d$ that record the birth and death of each topological feature. These diagrams are stable to small perturbations and can be compared via the bottleneck distance \citep{cohen2005stability}. Persistent homology thus provides a multiscale, noise-robust summary of the data’s intrinsic topological structure, forming the basis for the topology-preserving autoencoder. The core idea is to compute persistence diagrams for both data and latent spaces and penalize their topological difference, encouraging multiscale connectivity patterns to be maintained in the encoding. The topological loss is defined following \cite{moor2020topological} as:

\begin{align}
L_{X \to Z} &:= \tfrac{1}{2}\| A_X^{\pi_X} - A_Z^{\pi_X} \|^2, \label{eq:forward_consistency} \\
L_{Z \to X} &:= \tfrac{1}{2}\| A_Z^{\pi_Z} - A_X^{\pi_Z} \|^2. \label{eq:backward_consistency}
\end{align}

Here, $L_{X \to Z}$ and $L_{Z \to X}$ together form the topological loss 
$L_t = L_{X \to Z} + L_{Z \to X}$. For a mini-batch, \(X\), \(\hat{X}\), and \(Z\) denote the input, output, and latent spaces, respectively. 

Persistent homology is then computed on both $X$ and $Z$ to extract 
topologically relevant distances, with $\pi_X$ and $\pi_Z$ denoting the persistence pairings that indicate which distances in $A_X$ and $A_Z$ are topologically significant.
(i.e., indices of edges corresponding to topological features). Minimizing the combined objective Eq. \ref{eq:total_loss} enforces topological consistency, ensuring that the latent space preserves the structure of the data space. In the ideal case, when the latent and data space topologies are perfectly aligned, both $L_{X \to Z}$ and $L_{Z \to X}$ vanish.

\begin{align}
L &:= L_{\mathrm{rec}}(X, \hat{X}) + \lambda L_{t}. \label{eq:total_loss}
\end{align}

This term is added to the reconstruction loss of the autoencoder, where $\lambda$ controls the strength of the topological regularization. Gradients are computed efficiently since pairings change only under substantial perturbations, ensuring loss differentiability during training. The output of the TopoAE is considered a purified representation, where the latent space captures topology-preserving features derived solely from persistent homology (Fig.~\ref{fig:latent-comparison}). This process removes much of the adversarial perturbation in the input data, although the manifold reconstructed by the TopoAE is not fully aligned with classifier expectations. This motivates the role of the Reformer module in the pipeline. 

The overall training objective for the Reformer combines three losses: a mean squared error loss $\mathcal{L}_{\text{MSE}}$ for the pixel-level reconstruction between the TopoAE output and the VAE-reformed output, a cross-entropy loss $\mathcal{L}_{\text{CE}}$ between classifier predictions (logits) on the VAE-reformed output and the ground-truth labels, and a KL divergence term $\mathcal{L}_{\text{KL}}$ regularization. The MSE loss ensures the reconstruction remains structurally similar to the original image. In contrast, the cross-entropy loss encourages the Reformer to align the TopoAE output to the manifold expected by the classifier. The coefficients $\lambda_{1}$, $\lambda_{2}$, and $\lambda_{3}$ are the loss weights for each, respectively.

\begin{equation} \label{eq:total_loss_2}
\mathcal{L} = \lambda_{1}\,\mathcal{L}_{\text{MSE}} + \lambda_{2}\,\mathcal{L}_{\text{CE}} + \lambda_{3}\,\mathcal{L}_{\text{KL}} 
\end{equation}

The TopoReformer is trained separately on unperturbed samples from the training sets of MNIST \cite{lecun2010mnist}, EMNIST \cite{cohen2017emnistextensionmnisthandwritten}, and the OCR dataset \cite{belval2020textrecognitiondatagenerator} until convergence, following \cite{moor2020topological}. Afterward, its weights are frozen and used only for inference within the pipeline. The VAE and Auxiliary Module are then trained on the outputs of the TopoAE, namely, the latent vector and the topologically purified image, which are passed to the Auxiliary Module and VAE, respectively, using unperturbed inputs and the objective function in Eq.~\ref{eq:total_loss_2}. The classifier weights remain fixed throughout training.

\subsection{Auxiliary Training}

The Auxiliary Module receives the latent representation produced by the TopoAE via a learned latent projection network. This design leverages the fact that the TopoAE encodes inputs into a topology-aware latent space, where unperturbed and adversarially perturbed inputs occupy similar regions due to global manifold preservation. By feeding this representation to the Auxiliary Module, the Reformer can exploit this for more accurate predictions even under strong attacks.  The auxiliary supervision component, when introduced independently, exhibits limited effectiveness, yielding marginal or inconsistent improvements across attack settings. This behavior primarily arises from \textit{insufficient gradient propagation} to the auxiliary module, as the model tends to rely predominantly on the TopoReformer-purified output pathway rather than engaging with latent representations. A \textbf{Freeze-Flow training paradigm} is used to ensure proper training of the Auxiliary Module. Gradients are forcefully routed to it by first freezing the encoder of the Reformer VAE. After a set number of warmup epochs, the decoder is unfrozen, and the entire VAE and Auxiliary Module are trained jointly. The warmup phase ensures the auxiliary pathway receives stable gradients before the primary purified pathway dominates, effectively balancing optimization across both branches. This unified approach enables the auxiliary module to establish meaningful latent representations and improves robustness under stronger attacks. 

\subsection{Topological Latent Visualization}

\begin{figure}[ht]
    \centering
    \begin{subfigure}[b]{0.45\columnwidth}
        \centering
        \includegraphics[width=\textwidth]{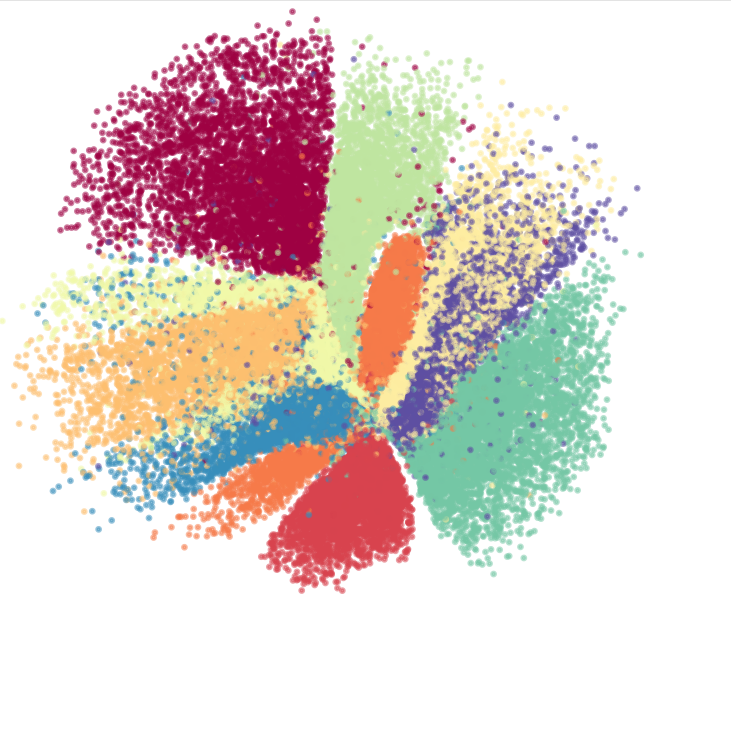}
        \caption{MNIST}
        \label{fig:latent1}
    \end{subfigure}
    \begin{subfigure}[b]{0.45\columnwidth}
        \centering
        \includegraphics[width=\textwidth]{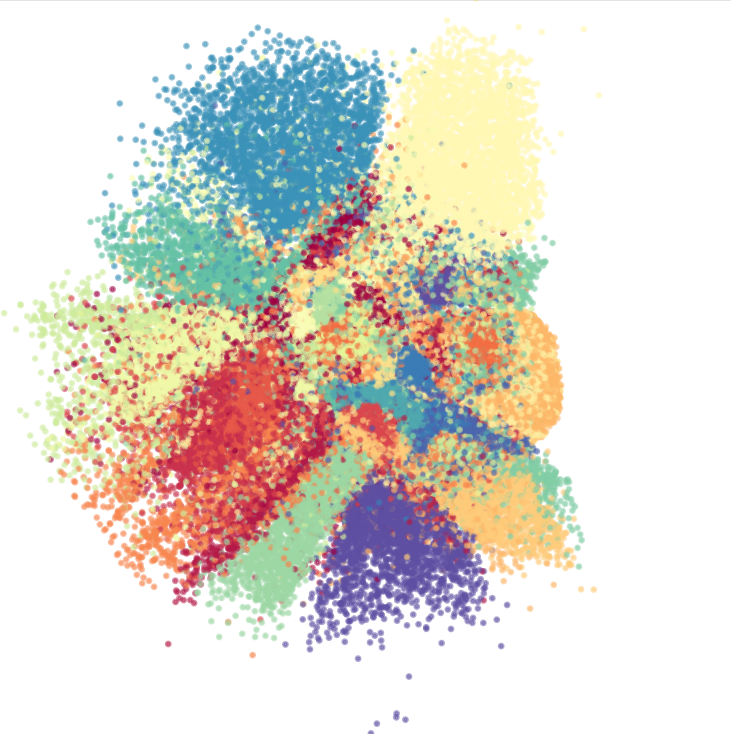}
        \caption{EMNIST}
        \label{fig:latent2}
    \end{subfigure}
    \caption{Comparison of topological latent visualizations.}
    \label{fig:latent-comparison}
\end{figure}

Fig. \ref{fig:latent-comparison} illustrates the motivation behind using topological autoencoders for adversarial and downstream OCR tasks. Even when trained with moderate noise additions, the TopoReformer produces consistent and separable latent space representations Eq. \eqref{eq:forward_consistency}, Eq. \eqref{eq:backward_consistency}. The autoencoder (AE) preserves digit class separation in the latent space, with each color cluster corresponding to a distinct digit or character. As observed in the MNIST latent map (Fig. ~\ref{fig:latent1}), the clusters are clean and radially separable, whereas in EMNIST (Fig. ~\ref{fig:latent2}), they appear more entangled due to the higher class diversity. However, the separation remains discernible even for highly imbalanced datasets.

\begin{table*}[t]
\caption{Ablation study under different  L$_\infty$ adversarial attacks on MNIST and EMNIST (reporting F1-score and Precision). Metrics are shown in percent (\%). Attack strengths are shown in the confidence parameter ($c$) and perturbation budget ($\epsilon$).}

\label{tab:ablation-attacks}
\begin{center}
\begin{tabular}{lll|cc|cc}
\multicolumn{3}{c}{} & \multicolumn{2}{c}{\bf MNIST} & \multicolumn{2}{c}{\bf EMNIST} 
\\ \hline \\
\multicolumn{1}{c}{\bf Classical Attacks} & \multicolumn{1}{c}{\bf Strength} & \multicolumn{1}{c}{\bf Ablation} & 
\multicolumn{1}{c}{\bf F1} & \multicolumn{1}{c}{\bf Precision} & 
\multicolumn{1}{c}{\bf F1} & \multicolumn{1}{c}{\bf Precision} 
\\ \hline \\

Carlini & Weak / Strong & No Defense & 30.41 / 4.30 & 31.23 / 6.78 & 36.54 / 33.85 & 53.59 / 50.99 \\
        & $c = 1e{-}2 / 1e{+}1$ & + TopoAE   & 53.92 / 48.51 & 61.35 / 50.68 & 30.87 / 27.71 & 66.00 / 64.10 \\
        &              & + Reformer & 65.38 / 67.93 & 71.11 / 69.83 & 50.64 / 49.16 & 59.71 / 58.29 \\
        &              & + Aux      & 65.36 / 72.41 & 73.17 / 74.85 & 64.98 / 63.92 & 69.37 / 68.50 \\
        &              & + Warmup   & \textbf{65.86 / 75.15} & \textbf{73.17 / 77.31} & \textbf{69.66 / 68.82} & \textbf{73.20 / 72.51} 
\\ \hline \\

PGD     & Weak / Strong & No Defense & 96.74 / 96.62 & 96.90 / 96.69 & 84.87 / 72.66  & 85.66 / 74.54 \\
        & $\epsilon = 0.005 / 0.01$ & + TopoAE   & 96.89 / 96.68 & 96.92 / 96.72 & 89.26 / 86.77 & 89.57 / 87.22 \\
        &              & + Reformer & 97.17 / 96.93 & 97.19 / 96.95 & \textbf{90.30 / 89.51} & \textbf{90.63 / 89.87} \\
        &              & + Aux      & \textbf{97.60 / 97.62} & \textbf{97.62 / 97.64} & 83.25 / 82.75 & 84.79 / 84.31 \\
        &              & + Warmup   & \textbf{97.70 / 97.62} & \textbf{97.72 / 97.64} & 84.53 / 83.79 & 85.12 / 84.43 
\\ \hline \\

FGSM    & Weak / Strong & No Defense & 96.87 / 96.61 & 96.89 / 96.65 & 90.83 / 72.59 & 91.04 / 73.83 \\
        & $\epsilon = 0.005 / 0.01$ & + TopoAE   & 96.86 / 96.63 & 96.90 / 96.89 & 90.48 / 85.47 & 90.73 / 86.04 \\
        &              & + Reformer & 97.01 / 96.94 & 97.04 / 96.96 & \textbf{90.95 / 89.17} & \textbf{91.14} / 84.94 \\
        &              & + Aux      & 97.60 / 97.21 & 97.62 / 97.62 & 90.49 / 83.60 & 90.73 / \textbf{89.41} \\
        &              & + Warmup   & \textbf{97.69 / 97.51} & \textbf{97.71 / 97.71} & 90.48 / 84.42 & 90.72 / 85.00 \\

\end{tabular}
\end{center}
\end{table*}

\section{Experimentation}

This section is divided into four parts: Datasets, OCR model attack setup, the classifiers used to evaluate performance, and visualization of latent features learned by TopoReformer. The classifiers are trained using the total loss function in Eq.~\ref{eq:total_loss} with loss weights $\lambda_1 = 1$, $\lambda_2 = 0.5$, and $\lambda_3 = 0.5$. Training is performed using Adam optimizer \cite{kingma2017adammethodstochasticoptimization} with a learning rate of $0.001$.

\subsection{Datasets}

The evaluation is conducted on three benchmark datasets, OCR Dataset \cite{belval2020textrecognitiondatagenerator, he2023protego}, MNIST \cite{lecun2010mnist}, and EMNIST \cite{cohen2017emnistextensionmnisthandwritten} letters. MNIST and EMNIST were chosen because their character sets closely resemble the inputs typically encountered by OCR models. Three classical adversarial attacks, PGD, FGSM, and Carlini, and two adaptive adversarial attacks, EOT  and BPDA, are applied to the test sets, with classical attacks evaluated at both strong and weak strength levels. Comparisons are made against a self-constructed baseline, as existing benchmarks for classification under adversarial purification with topology-preserving autoencoders are limited. Strength levels for PGD and FGSM are selected as done in \cite{meng2017magnet}. These works report performance on adversarial attack detection rather than classification tasks. For Carlini and other adaptive attacks, the strength is adjusted until classifier performance degrades, establishing a baseline on which components are added to demonstrate incremental improvements. The experimentation is extended to attacks specific to OCR models.The OCR training set was generated using the open-source code available here \cite{belval2020textrecognitiondatagenerator}. The dataset spans 53 classes, including uppercase and lowercase characters and a blank token. Unlike MNIST and EMNIST, this dataset is word-based, introducing additional structural challenges necessitating character-wise processing. 

\begin{table*}[ht]
\caption{Ablation study under adaptive adversarial attacks on MNIST and EMNIST datasets (reporting Attack Success Rate, F1-score, and Precision). Metrics are shown in percent (\%).}
\label{tab:ocr-attacks}
\begin{center}

\renewcommand{\arraystretch}{1.3}
\setlength{\tabcolsep}{8pt}


\begin{tabular}{ll|ccc|ccc}
\multicolumn{2}{c}{} & \multicolumn{3}{c}{\bf{MNIST}} & \multicolumn{3}{c}{\bf{EMNIST}} \\
\hline
\multicolumn{1}{c}{\textbf{Adaptive Attack}} &
\multicolumn{1}{c}{\textbf{Defense}} &
\multicolumn{1}{c}{\textbf{ASR $\downarrow$}} &
\multicolumn{1}{c}{\textbf{F1 $\uparrow$}} &
\multicolumn{1}{c}{\textbf{Precision $\uparrow$}}&
\multicolumn{1}{c}{\textbf{ASR $\downarrow$}} &
\multicolumn{1}{c}{\textbf{F1 $\uparrow$}} &
\multicolumn{1}{c}{\textbf{Precision $\uparrow$}} 
\\ \hline \\

EOT & No Defense & 99.05 & 3.38 & 5.21& 97.73 & 1.42 & 2.67 \\
    & TopoReformer     & \textbf{9.19} & \textbf{90.73} & \textbf{91.07}& \textbf{28.32} & \textbf{73.28} & \textbf{75.12} \\[4pt]
\hline

EOT+BPDA & No Defense & 99.70 & 3.21 & 4.89& 98.69& 1.36 & 2.54 \\
         & TopoReformer     & 36.59 & 64.71 & 66.65&44.26 & 58.92 & 61.31 \\[4pt]
\hline

BPDA & No Defense & 99.60 & 4.44 & 7.28& 95.30 & 2.03 & 5.17 \\
     & TopoReformer     & 81.14 & 15.65 & 40.98& 84.46 & 12.77 & 35.42 \\

\end{tabular}
\end{center}
\end{table*}

\subsection{OCR Model Adversarial Attack Setup}

FAWA \cite{chen2020fawa} is a white-box, targeted attack tailored for OCR that embeds adversarial signals within plausible “watermark” overlays. By constraining perturbations to visually natural overlays commonly seen in documents, FAWA achieves near-perfect attack success with substantially reduced perturbation magnitude and iterations compared to pixel-space baselines. Evaluations are shown using character-level performance and attack success rate (ASR), where a 100\% ASR indicates that all adversarial images are misidentified as the targeted texts by the threat model. Table \ref{tab:ocr-ablation} summarizes the results. However, applying this directly to OCR words proved ineffective, as word-level images lack a consistent geometric structure. To address this, words were decomposed into individual characters. An auto-padding step was applied to ensure that all decomposed characters had a uniform size, as required by the model. Later, each character was centered with auto-padding to dimensions $2 \times 28 \times 44$. Following processing and purification at the character level, the characters were recombined into their original word-level format of $32 \times 100$. This design allowed the pipeline to exploit character-level features even in word-based OCR, thereby preserving robustness and accuracy. 

\subsection{Classifier Architectures}

The MNIST and EMNIST classifiers in the pipeline use the architectures as described in the Magnet \cite{meng2017magnet}. The OCR architectures used in this work encompass CTC-based and attention-based designs to provide a broad evaluation of the proposed defense. The three CTC-based models CRNN \cite{shi2016end}, Rosetta \cite{borisyuk2018rosetta}, and STAR-Net \cite{liu2016star} combine convolutional feature extraction with sequence modeling and CTC-based decoding, differing in backbone complexity and feature aggregation. The two attention-based models RARE \cite{shi2016robust} and TRBA \cite{baek2019wrong} employ spatial attention and cross-entropy-based decoding to capture context and sequential dependencies, representing a complementary approach to CTC frameworks. Together, these architectures cover a representative spectrum of modern OCR pipelines.

\section{Results}

The proposed method is evaluated using quantitative and qualitative analyses to assess its robustness and interpretability. Quantitative results include ablation studies that isolate the contribution of each component. OCR-specific evaluations demonstrate the transferability of the defense to sequence-based recognition tasks. Qualitative analysis is through Grad-CAM visualizations~\cite{selvaraju2017grad}.

\subsection{Quantitative Results}

\subsubsection{Classical Attacks Results} Table \ref{tab:ablation-attacks} presents an ablation study evaluating the contributions of each architectural component under three adversarial attack scenarios on MNIST and EMNIST. Strength levels for PGD and FGSM follow the approach in \cite{meng2017magnet}, while for the Carlini–Wagner (C\&W) attack, the perturbation is adjusted until the classifier performs poorly. The standalone classifier exhibits significant vulnerability, achieving F1 scores of 30.41\% on MNIST and 36.54\% on EMNIST under weak C\&W attacks. Incorporating TopoReformer substantially improves performance, raising F1 scores to 65.86\% on MNIST and 69.66\% on EMNIST. The demonstration of the complementary effect of the freeze flow training paradigm is also clearly seen; any anomaly in the ablations can be attributed to the manifold mismatch between the TopoReformer output and the classifier's inputs. C\&W generates minimal, high-confidence perturbations that reduce baseline accuracy to near zero \cite{carlini2017towards}, leaving considerable room for corrective gains. In contrast, PGD and FGSM are first-order attacks that do not always succeed, resulting in higher baseline performance and correspondingly smaller absolute improvements. Prior studies have shown that iterative PGD and multi-step FGSM attacks often saturate on standard datasets \cite{madry2017towards, meng2017magnet}, limiting the measurable effect of purification. Furthermore, TopoReformer enforces global manifold coherence rather than explicitly regularizing local gradients, making it particularly effective against fragile, low-distortion C\&W perturbations while producing moderate gains for already strong FGSM and PGD attacks \cite{goodfellow2014explaining, tramer2020adaptive} as seen above.

\begin{table*}[ht]
\caption{Ablation study of OCR-specific FAWA adversarial attack on diverse OCR Models. Metrics include Attack Success Rate (ASR, lower is better) and character-level performance (higher is better). All metrics are shown in percent (\%).}
\label{tab:ocr-ablation}
\begin{center}

\renewcommand{\arraystretch}{1.3}  
\setlength{\tabcolsep}{10pt}       

\begin{tabular}{lllccc}
\hline
\textbf{Model Type} & \textbf{Model} & \textbf{Defense} & \textbf{ASR $\downarrow$} & \textbf{Char Accuracy $\uparrow$} & \textbf{Char Precision $\uparrow$} \\
\hline

\textbf{CTC-based} & CRNN  & No Defense & 100 & 48.13 & 45.69 \\
                   &               & Defense     & 78.83 & 71.00 & 71.61 \\
\cline{2-6}
                   & Rosetta  & No Defense & 99.83 & 69.66 & 62.94 \\
                   &                 & Defense     & 44.08 & 85.98 & 87.52 \\
\cline{2-6}
                   & STAR-Net  & No Defense & 98.92 & 74.52 & 69.77 \\
                   &                  & Defense     & 65.17 & 79.81 & 81.44 \\
\hline

\textbf{Attention-based} & RARE  & No Defense & 99.92 & 51.25 & 49.77 \\
                         &               & Defense     & 87.67 & 65.18 & 64.74 \\
\cline{2-6}
                         & TRBA  & No Defense & 99.83 & 46.68 & 44.26 \\
                         &              & Defense     & 60.75 & 80.26 & 80.81 \\
\hline
\end{tabular}

\end{center}
\end{table*}

\begin{figure*}[t]
\centering
\scriptsize
\begin{tabular}{cccc}
    \includegraphics[width=0.15\textwidth]{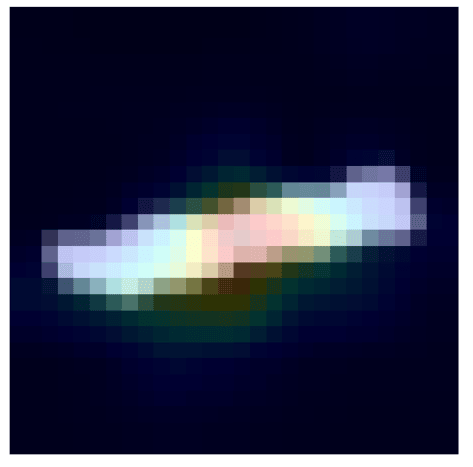} &
    \includegraphics[width=0.15\textwidth]{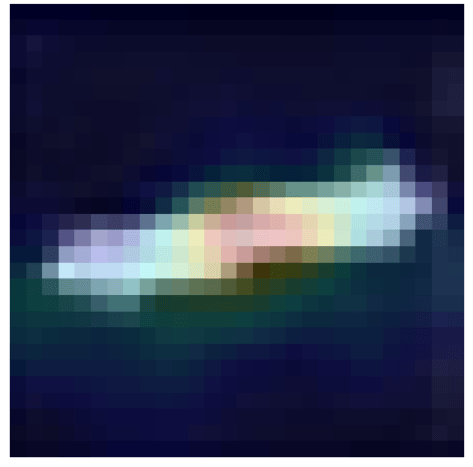} &
    \includegraphics[width=0.15\textwidth]{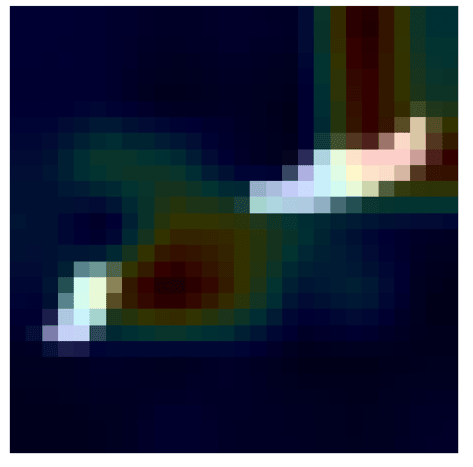} &
    \includegraphics[width=0.15\textwidth]{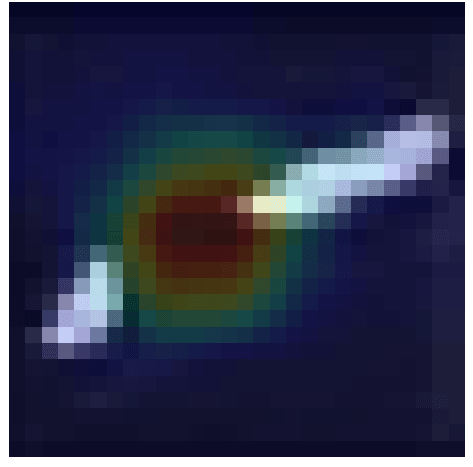} \\
    \parbox{0.15\textwidth}{\centering (a) Unperturbed input\\label: i, pred: l (0.38)} &
    \parbox{0.15\textwidth}{\centering (b) Reformed output\\label: i, pred: i (0.51)} &
    \parbox{0.15\textwidth}{\centering (c) Adversarial input\\label: i, pred: j (0.18)} &
    \parbox{0.15\textwidth}{\centering (d) Reformed output\\label: i, pred: i (0.88)} \\
\end{tabular}
\caption{Comparison of Grad-CAM visualizations. (a)–(b) Unperturbed images, (c)–(d) adversarial images. Reformers project inputs toward topology-consistent manifolds, leading to more focused and confident class-discriminative regions.}
\label{fig:gradcam-emnist}
\end{figure*}

\subsubsection{Adaptive Attack Results}
To avoid a false sense of security \cite{athalye2018obfuscated} Table \ref{tab:ocr-attacks} shows TopoReformer’s behavior is analyzed under three adaptive settings (EOT, BPDA, EOT+BPDA). The model's strong performance, particularly under EOT, indicates that its defense mechanism is not merely denoising or gradient obfuscation. \cite{tramer2020adaptive} emphasized that robust evaluation should expose not just whether a defense breaks but also how different adaptive strategies interact. Following this philosophy, while BPDA reduces the defense performance, EOT-based attacks are less effective, revealing how topological constraints reshape gradient landscapes. The proposed method integrates a topological autoencoder constraint into the latent space and achieves notable robustness against standard and adaptive attacks such as FGSM, CW, and EOT. Interestingly, this robustness arises without explicit gradient regularization, suggesting that preserving global manifold topology implicitly constrains the model’s \textbf{Lipschitz continuity} \cite{tsuzuku2018lipschitz}, that is, it limits how abruptly latent representations can vary in response to input perturbations. By maintaining the continuity and connectivity of the data manifold, the model resists large deformations in representation space, preserving class separation even under stochastic or expectation-based attacks. The model retains strong performance under EOT and EOT+BPDA attacks, indicating that the topological constraint promotes global stability and smooth transitions across the latent manifold. However, under BPDA alone, the defense weakens, consistent with the hypothesis that the topological regularizer enforces global Lipschitz-like smoothness but does not directly constrain fine-grained local curvature, which BPDA explicitly exploits. Overall, these results highlight that topology-based regularization enhances global geometric consistency, offering a complementary axis of robustness that operates at the manifold level rather than through explicit gradient control. Furthermore, the classifiers maintain approximately 98\% accuracy on MNIST and 94\% on EMNIST in unperturbed (clean ) settings. The negligible variation in unperturbed performance confirms that the unified TopoReformer pipeline enhances adversarial robustness without compromising standard classification capability.

\subsubsection{OCR Ablation Results}
Table~\ref{tab:ocr-ablation} summarizes the OCR-specific adversarial defense evaluation conducted across five representative architectures: three CTC-based models, CRNN, Rosetta, and STAR-Net using CTC loss, and two attention-based models, RARE and TRBA, using cross-entropy loss. Each model is evaluated under two configurations: the baseline \textit{No Defense} setup and the proposed \textit{TopoReformer} configuration. Baseline OCR models demonstrate complete vulnerability to adversarial perturbations, with ASR values near 100\% and considerable drops in character accuracy (e.g., 48.13\% for CRNN). Incorporating the TopoReformer module substantially enhances robustness across all architectures, consistently reducing ASR and improving accuracy, precision, and recall. For instance, in the CRNN model, the ASR drops to 78.83\% while accuracy increases to 71\%, and similar gains are observed for the other models. Even though adding the Reformer component yielded notable improvements in MNIST and EMNIST experiments, it was intentionally excluded from OCR evaluations to preserve deployability efficiency.

\subsection{Qualitative Results}

To interpret the effect of the topological constraint, Grad-CAM is used \cite{selvaraju2017grad} to visualize how the model’s attention changes under adversarial perturbations. Grad-CAM produces a spatial heatmap highlighting the most influential regions for prediction. Fig. \ref{fig:gradcam-emnist} compares the baselines and TopoReformer's attention maps. The baseline shows scattered or shifted activations under adversarial noise, often focusing on irrelevant regions. TopoReformer maintains compact, semantically consistent activations, concentrating on the object of interest even under perturbation. Another noteworthy observation is that the confidence scores of unperturbed images also increase after being processed through the TopoAE. This is particularly interesting, as most adversarial training techniques tend to improve robustness at the expense of performance on unperturbed samples. These results support the hypothesis that enforcing topological consistency in the latent manifold encourages globally coherent and stable attention, reducing representational drift. 

\section{Conclusion}

This work demonstrates that topological feature analysis and persistent homology–based purification form a viable defense pipeline. Experimental results on MNIST and EMNIST demonstrate that TopoReformer substantially outperforms baseline classifiers under standard adversarial attacks such as FGSM, PGD, and C\&W. It also remains effective against adaptive attacks like EOT and EOT+BPDA, as well as OCR-specific threats, including FAWA watermark-based perturbations. By leveraging the inherent stability of topological representations, TopoReformer preserves the global structure of the data manifold, naturally filtering out adversarial noise and maintaining semantic integrity. These results consistently reduce attack success rates and improve robustness across all evaluated metrics, highlighting the effectiveness of topology-driven purification under both standard and adaptive adversarial settings.

\section{Future Work}

Future work will focus on integrating topology-aware training directly into OCR model optimization, thereby removing the need for a separate Reformer stage. Another direction involves constructing a comprehensive benchmark dataset and standardized classifier suite for the evaluation of OCR-specific adversarial attacks and solutions. This benchmark will encompass both classical and adaptive threat models, addressing the current gap in publicly available resources for OCR robustness assessment.

\bibliography{aaai2026}

@inproceedings{liu2016star,
  title={Star-net: a spatial attention residue network for scene text recognition.},
  author={Liu, Wei and Chen, Chaofeng and Wong, Kwan-Yee K and Su, Zhizhong and Han, Junyu},
  booktitle={BMVC},
  volume={2},
  pages={7},
  year={2016}
}

@article{tramer2020adaptive,
  title={On adaptive attacks to adversarial example defenses},
  author={Tramer, Florian and Carlini, Nicholas and Brendel, Wieland and Madry, Aleksander},
  journal={Advances in neural information processing systems},
  volume={33},
  pages={1633--1645},
  year={2020}
}

@article{shi2016end,
  title={An end-to-end trainable neural network for image-based sequence recognition and its application to scene text recognition},
  author={Shi, Baoguang and Bai, Xiang and Yao, Cong},
  journal={IEEE transactions on pattern analysis and machine intelligence},
  volume={39},
  number={11},
  pages={2298--2304},
  year={2016},
  publisher={IEEE}
}

@inproceedings{athalye2018obfuscated,
  title={Obfuscated gradients give a false sense of security: Circumventing defenses to adversarial examples},
  author={Athalye, Anish and Carlini, Nicholas and Wagner, David},
  booktitle={International conference on machine learning},
  pages={274--283},
  year={2018},
  organization={PMLR}
}

@inproceedings{baek2019wrong,
  title={What is wrong with scene text recognition model comparisons? dataset and model analysis},
  author={Baek, Jeonghun and Kim, Geewook and Lee, Junyeop and Park, Sungrae and Han, Dongyoon and Yun, Sangdoo and Oh, Seong Joon and Lee, Hwalsuk},
  booktitle={Proceedings of the IEEE/CVF international conference on computer vision},
  pages={4715--4723},
  year={2019}
}

@inproceedings{shi2016robust,
  title={Robust scene text recognition with automatic rectification},
  author={Shi, Baoguang and Wang, Xinggang and Lyu, Pengyuan and Yao, Cong and Bai, Xiang},
  booktitle={Proceedings of the IEEE conference on computer vision and pattern recognition},
  pages={4168--4176},
  year={2016}
}

@inproceedings{borisyuk2018rosetta,
  title={Rosetta: Large scale system for text detection and recognition in images},
  author={Borisyuk, Fedor and Gordo, Albert and Sivakumar, Viswanath},
  booktitle={Proceedings of the 24th ACM SIGKDD international conference on knowledge discovery \& data mining},
  pages={71--79},
  year={2018}
}

@inproceedings{bazzo2020assessing,
  title={Assessing the impact of OCR errors in information retrieval},
  author={Bazzo, Guilherme Torresan and Lorentz, Gustavo Acauan and Suarez Vargas, Danny and Moreira, Viviane P},
  booktitle={European Conference on Information Retrieval},
  pages={102--109},
  year={2020},
  organization={Springer}
}

@article{horn2021topological,
  title={Topological graph neural networks},
  author={Horn, Max and De Brouwer, Edward and Moor, Michael and Moreau, Yves and Rieck, Bastian and Borgwardt, Karsten},
  journal={arXiv preprint arXiv:2102.07835},
  year={2021}
}

@article{hwang2019puvae,
  title={Puvae: A variational autoencoder to purify adversarial examples},
  author={Hwang, Uiwon and Park, Jaewoo and Jang, Hyemi and Yoon, Sungroh and Cho, Nam Ik},
  journal={IEEE Access},
  volume={7},
  pages={126582--126593},
  year={2019},
  publisher={IEEE}
}

@article{sainburg2021parametric,
  title={Parametric UMAP embeddings for representation and semisupervised learning},
  author={Sainburg, Tim and McInnes, Leland and Gentner, Timothy Q},
  journal={Neural Computation},
  volume={33},
  number={11},
  pages={2881--2907},
  year={2021},
  publisher={MIT Press One Rogers Street, Cambridge, MA 02142-1209, USA journals-info~…}
}

@misc{ngairangbam2025enhancinganomalydetectiontopologyaware,
      title={Enhancing anomaly detection with topology-aware autoencoders}, 
      author={Vishal S. Ngairangbam and Błażej Rozwoda and Kazuki Sakurai and Michael Spannowsky},
      year={2025},
      eprint={2502.10163},
      archivePrefix={arXiv},
      primaryClass={hep-ph},
      url={https://arxiv.org/abs/2502.10163}, 
}

@article{karthikeyan2021ocr,
  title={An OCR post-correction approach using deep learning for processing medical reports},
  author={Karthikeyan, Srinidhi and de Herrera, Alba G Seco and Doctor, Faiyaz and Mirza, Asim},
  journal={IEEE Transactions on Circuits and Systems for Video Technology},
  volume={32},
  number={5},
  pages={2574--2581},
  year={2021},
  publisher={IEEE}
}

@article{pingili2025ai,
  title={AI-driven intelligent document processing for banking and finance},
  author={Pingili, Ramesh},
  journal={International Journal of Management \& Entrepreneurship Research},
  volume={7},
  number={2},
  pages={98--109},
  year={2025}
}

@inproceedings{kakani2017improved,
  title={Improved OCR based automatic vehicle number plate recognition using features trained neural network},
  author={Kakani, Bhavin V and Gandhi, Divyang and Jani, Sagar},
  booktitle={2017 8th international conference on computing, communication and networking technologies (ICCCNT)},
  pages={1--6},
  year={2017},
  organization={IEEE}
}

@article{li2022review,
  title={A review of adversarial attack and defense for classification methods},
  author={Li, Yao and Cheng, Minhao and Hsieh, Cho-Jui and Lee, Thomas CM},
  journal={The American Statistician},
  volume={76},
  number={4},
  pages={329--345},
  year={2022},
  publisher={Taylor \& Francis}
}

@article{goodfellow2014explaining,
  title={Explaining and harnessing adversarial examples},
  author={Goodfellow, Ian J and Shlens, Jonathon and Szegedy, Christian},
  journal={arXiv preprint arXiv:1412.6572},
  year={2014}
}

@inproceedings{carlini2017towards,
  title={Towards evaluating the robustness of neural networks},
  author={Carlini, Nicholas and Wagner, David},
  booktitle={2017 ieee symposium on security and privacy (sp)},
  pages={39--57},
  year={2017},
  organization={Ieee}
}

@article{madry2017towards,
  title={Towards deep learning models resistant to adversarial attacks},
  author={Madry, Aleksander and Makelov, Aleksandar and Schmidt, Ludwig and Tsipras, Dimitris and Vladu, Adrian},
  journal={arXiv preprint arXiv:1706.06083},
  year={2017}
}

@inproceedings{chen2020fawa,
  title={FAWA: Fast adversarial watermark attack on optical character recognition (OCR) systems},
  author={Chen, Lu and Sun, Jiao and Xu, Wei},
  booktitle={Joint European Conference on Machine Learning and Knowledge Discovery in Databases},
  pages={547--563},
  year={2020},
  organization={Springer}
}

@inproceedings{moor2020topological,
  title={Topological autoencoders},
  author={Moor, Michael and Horn, Max and Rieck, Bastian and Borgwardt, Karsten},
  booktitle={International conference on machine learning},
  pages={7045--7054},
  year={2020},
  organization={PMLR}
}

@inproceedings{folz2020adversarial,
  title={Adversarial defense based on structure-to-signal autoencoders},
  author={Folz, Joachim and Palacio, Sebastian and Hees, Joern and Dengel, Andreas},
  booktitle={2020 IEEE Winter Conference on Applications of Computer Vision (WACV)},
  pages={3568--3577},
  year={2020},
  organization={IEEE}
}

@inproceedings{he2023protego,
  title={ProTegO: Protect text content against OCR extraction attack},
  author={He, Yanru and Chen, Kejiang and Chen, Guoqiang and Ma, Zehua and Zhang, Kui and Zhang, Jie and Bian, Huanyu and Fang, Han and Zhang, Weiming and Yu, Nenghai},
  booktitle={Proceedings of the 31st ACM International Conference on Multimedia},
  pages={7424--7434},
  year={2023}
}

@article{vu2025topological,
  title={Topological Signatures of Adversaries in Multimodal Alignments},
  author={Vu, Minh and Zollicoffer, Geigh and Mai, Huy and Nebgen, Ben and Alexandrov, Boian and Bhattarai, Manish},
  journal={arXiv preprint arXiv:2501.18006},
  year={2025}
}

@article{kuang2024defense,
  title={Defense against adversarial attacks using topology aligning adversarial training},
  author={Kuang, Huafeng and Liu, Hong and Lin, Xianming and Ji, Rongrong},
  journal={IEEE Transactions on Information Forensics and Security},
  volume={19},
  pages={3659--3673},
  year={2024},
  publisher={IEEE}
}

@inproceedings{athalye2018synthesizing,
  title={Synthesizing robust adversarial examples},
  author={Athalye, Anish and Engstrom, Logan and Ilyas, Andrew and Kwok, Kevin},
  booktitle={International conference on machine learning},
  pages={284--293},
  year={2018},
  organization={PMLR}
}

@inproceedings{cohen2005stability,
  title={Stability of persistence diagrams},
  author={Cohen-Steiner, David and Edelsbrunner, Herbert and Harer, John},
  booktitle={Proceedings of the twenty-first annual symposium on Computational geometry},
  pages={263--271},
  year={2005}
}

@article{tsuzuku2018lipschitz,
  title={Lipschitz-margin training: Scalable certification of perturbation invariance for deep neural networks},
  author={Tsuzuku, Yusuke and Sato, Issei and Sugiyama, Masashi},
  journal={Advances in neural information processing systems},
  volume={31},
  year={2018}
}

@article{gebhart2017adversary,
  title={Adversary detection in neural networks via persistent homology},
  author={Gebhart, Thomas and Schrater, Paul},
  journal={arXiv preprint arXiv:1711.10056},
  year={2017}
}

@article{goibert2022adversarial,
  title={An adversarial robustness perspective on the topology of neural networks},
  author={Goibert, Morgane and Ricatte, Thomas and Dohmatob, Elvis},
  journal={arXiv preprint arXiv:2211.02675},
  year={2022}
}

@article{guo2017countering,
  title={Countering adversarial images using input transformations},
  author={Guo, Chuan and Rana, Mayank and Cisse, Moustapha and Van Der Maaten, Laurens},
  journal={arXiv preprint arXiv:1711.00117},
  year={2017}
}

@article{dhillon2018stochastic,
  title={Stochastic activation pruning for robust adversarial defense},
  author={Dhillon, Guneet S and Azizzadenesheli, Kamyar and Lipton, Zachary C and Bernstein, Jeremy and Kossaifi, Jean and Khanna, Aran and Anandkumar, Anima},
  journal={arXiv preprint arXiv:1803.01442},
  year={2018}
}

@article{chen2025adversarial,
  title={Adversarial Masked Autoencoder Purifier with Defense Transferability},
  author={Chen, Yuan-Chih and Lu, Chun-Shien},
  journal={arXiv preprint arXiv:2501.16904},
  year={2025}
}

@inproceedings{yang2024adversarial,
  title={Adversarial purification with the manifold hypothesis},
  author={Yang, Zhaoyuan and Xu, Zhiwei and Zhang, Jing and Hartley, Richard and Tu, Peter},
  booktitle={Proceedings of the AAAI Conference on Artificial Intelligence},
  volume={38},
  pages={16379--16387},
  year={2024}
}

@article{kalaria2022towards,
  title={Towards adversarial purification using denoising autoencoders},
  author={Kalaria, Dvij and Hazra, Aritra and Chakrabarti, Partha Pratim},
  journal={arXiv preprint arXiv:2208.13838},
  year={2022}
}

@inproceedings{meng2017magnet,
  title={Magnet: a two-pronged defense against adversarial examples},
  author={Meng, Dongyu and Chen, Hao},
  booktitle={Proceedings of the 2017 ACM SIGSAC conference on computer and communications security},
  pages={135--147},
  year={2017}
}

@article{zhao2022adversarial,
  title={Adversarial training methods for deep learning: A systematic review},
  author={Zhao, Weimin and Alwidian, Sanaa and Mahmoud, Qusay H},
  journal={Algorithms},
  volume={15},
  number={8},
  pages={283},
  year={2022},
  publisher={MDPI}
}

@article{he2022your,
  title={Be your own neighborhood: Detecting adversarial example by the neighborhood relations built on self-supervised learning},
  author={He, Zhiyuan and Yang, Yijun and Chen, Pin-Yu and Xu, Qiang and Ho, Tsung-Yi},
  journal={arXiv preprint arXiv:2209.00005},
  year={2022}
}

@article{edelsbrunner2008persistent,
  title={Persistent homology-a survey},
  author={Edelsbrunner, Herbert and Harer, John and others},
  journal={Contemporary mathematics},
  volume={453},
  number={26},
  pages={257--282},
  year={2008},
  publisher={Providence, RI: American Mathematical Society}
}

@article{lecun2010mnist,
         title={MNIST handwritten digit database},
         author={LeCun, Yann and Cortes, Corinna and Burges, CJ},
         journal={ATT Labs [Online]. Available: http://yann.lecun.com/exdb/mnist},
         volume={2},
         year={2010}
}

@misc{cohen2017emnistextensionmnisthandwritten,
      title={EMNIST: an extension of MNIST to handwritten letters}, 
      author={Gregory Cohen and Saeed Afshar and Jonathan Tapson and André van Schaik},
      year={2017},
      eprint={1702.05373},
      archivePrefix={arXiv},
      primaryClass={cs.CV},
      url={https://arxiv.org/abs/1702.05373}, 
}

@article{akhtar2018threat,
  title={Threat of adversarial attacks on deep learning in computer vision: A survey},
  author={Akhtar, Naveed and Mian, Ajmal},
  journal={Ieee Access},
  volume={6},
  pages={14410--14430},
  year={2018},
  publisher={IEEE}
}

@article{chakraborty2021survey,
  title={A survey on adversarial attacks and defences},
  author={Chakraborty, Anirban and Alam, Manaar and Dey, Vishal and Chattopadhyay, Anupam and Mukhopadhyay, Debdeep},
  journal={CAAI Transactions on Intelligence Technology},
  volume={6},
  number={1},
  pages={25--45},
  year={2021},
  publisher={Wiley Online Library}
}

@inproceedings{zomorodian2004computing,
  title={Computing persistent homology},
  author={Zomorodian, Afra and Carlsson, Gunnar},
  booktitle={Proceedings of the twentieth annual symposium on Computational geometry},
  pages={347--356},
  year={2004}
}

@inproceedings{cintas2021detecting,
  title={Detecting adversarial attacks via subset scanning of autoencoder activations and reconstruction error},
  author={Cintas, Celia and Speakman, Skyler and Akinwande, Victor and Ogallo, William and Weldemariam, Komminist and Sridharan, Srihari and McFowland, Edward},
  booktitle={Proceedings of the twenty-ninth international conference on international joint conferences on artificial intelligence},
  pages={876--882},
  year={2021}
}

@article{qian2022survey,
  title={A survey of robust adversarial training in pattern recognition: Fundamental, theory, and methodologies},
  author={Qian, Zhuang and Huang, Kaizhu and Wang, Qiu-Feng and Zhang, Xu-Yao},
  journal={Pattern Recognition},
  volume={131},
  pages={108889},
  year={2022},
  publisher={Elsevier}
}

@inproceedings{lu2018limitation,
  title={On the limitation of MagNet defense against L1-based adversarial examples},
  author={Lu, Pei-Hsuan and Chen, Pin-Yu and Chen, Kang-Cheng and Yu, Chia-Mu},
  booktitle={2018 48th Annual IEEE/IFIP International Conference on Dependable Systems and Networks Workshops (DSN-W)},
  pages={200--214},
  year={2018},
  organization={IEEE}
}

@article{imam2022ocr,
  title={OCR post-correction for detecting adversarial text images},
  author={Imam, Niddal H and Vassilakis, Vassilios G and Kolovos, Dimitris},
  journal={Journal of Information Security and Applications},
  volume={66},
  pages={103170},
  year={2022},
  publisher={Elsevier}
}

@article{xie2017mitigating,
  title={Mitigating adversarial effects through randomization},
  author={Xie, Cihang and Wang, Jianyu and Zhang, Zhishuai and Ren, Zhou and Yuille, Alan},
  journal={arXiv preprint arXiv:1711.01991},
  year={2017}
}

@inproceedings{selvaraju2017grad,
  title={Grad-cam: Visual explanations from deep networks via gradient-based localization},
  author={Selvaraju, Ramprasaath R and Cogswell, Michael and Das, Abhishek and Vedantam, Ramakrishna and Parikh, Devi and Batra, Dhruv},
  booktitle={Proceedings of the IEEE international conference on computer vision},
  pages={618--626},
  year={2017}
}

@article{bakhti2019ddsa,
  title={DDSA: A defense against adversarial attacks using deep denoising sparse autoencoder},
  author={Bakhti, Yassine and Fezza, Sid Ahmed and Hamidouche, Wassim and D{\'e}forges, Olivier},
  journal={IEEE Access},
  volume={7},
  pages={160397--160407},
  year={2019},
  publisher={IEEE}
}

@misc{belval2020textrecognitiondatagenerator,
  author       = {Belval},
  title        = {{TextRecognitionDataGenerator}},
  year         = {2020},
  howpublished = {\url{https://github.com/Belval/TextRecognitionDataGenerator}},
  note         = {Accessed: 30 August 2025}
}

@misc{kingma2017adammethodstochasticoptimization,
      title={Adam: A Method for Stochastic Optimization}, 
      author={Diederik P. Kingma and Jimmy Ba},
      year={2017},
      eprint={1412.6980},
      archivePrefix={arXiv},
      primaryClass={cs.LG},
      url={https://arxiv.org/abs/1412.6980}, 
}

\end{document}